\documentclass{article}
\usepackage{spconf,amsmath,graphicx,subcaption,adjustbox}

\title{SALIENCY-ENHANCED ROBUST VISUAL TRACKING}
\name{Caglar Aytekin, Francesco Cricri and Emre Aksu}
\address{Nokia Technologies, Tampere, Finland}
\begin{document}
%
\maketitle
\begin{abstract}
Discrete correlation filter (DCF) based trackers have shown considerable success in visual object tracking. These trackers often make use of low to mid level features such as histogram of gradients (HoG) and mid-layer activations from convolution neural networks (CNNs). We argue that including semantically higher level information to the tracked features may provide further robustness to challenging cases such as viewpoint changes. Deep salient object detection is one example of such high level features, as it make use of semantic information to highlight the important regions in the given scene. In this work, we propose an improvement over DCF based trackers by combining saliency based and other features based filter responses. This combination is performed with an adaptive weight on the saliency based filter responses, which is automatically selected according to the temporal consistency of visual saliency. We show that our method consistently improves a baseline DCF based tracker especially in challenging cases and performs superior to the state-of-the-art. Our improved tracker operates at 9.3 fps, introducing a small computational burden over the baseline which operates at 11 fps.
\end{abstract}
\section{Introduction}
\label{sec:intro}

The aim of visual tracking is to successfully locate an object over time throughout a video. 
The object's initial location and spatial extent is provided when the object appears, either manually or by automatic detection. 
There are many approaches to visual tracking such as particle filters \cite{parfil}, tracking by detection \cite{tld} and discriminative correlation filter (DCF) based \cite{eco}, \cite{mosse}, \cite{ccot} approaches. 
Recent DCF based tracking methods have shown state-of-the-art performance \cite{eco}, \cite{ccot}. DCF based trackers generally have three basic steps. First, a correlation filter is learned from the appearance of the object at a given frame. Second, the object is searched with the learned filter in the next frame. Finally filter is updated according to the new appearance which is obtained from the location found in the second step. 

Successful DCF based trackers use low and mid-level features.
For example ECO \cite{eco} uses histogram of gradients \cite{hog} (HoG) and the features extracted by conv1 and conv5 layers of pre-trained VGG-m network \cite{vgg}.
We argue that semantically higher level information such as pixel-wise predictions are intuitively more robust to appearance and viewpoint changes.
For example, a successful person detector will always give consistent (1 on the person, 0 elsewhere) results regardless of appearance and viewpoint changes, whereas the features that are used in person detection before the classification step may vary with these changes.

One potential high-level information can be semantic segmentation predictions \cite{fcn}, \cite{dlab}, however accurate semantic segmentation usually comes at a high computational cost \cite{fcn}, \cite{dlab}.
This is not desired since one may wish to keep the near real-time speed of the DCF based trackers \cite{eco}.
Another pixel-wise classification task is salient object segmentation \cite{salobj}.
With the recent advent in deep salient object segmentation, methods such as \cite{dhsnet} can produce very accurate saliency maps with near real-time run times.
Due to high accuracy and fast processing times, in this paper we propose to incorporate saliency as a feature to be tracked into DCF based trackers.

Exploiting visual saliency in tracking has been a topic of other studies as well.
In \cite{sdisc}, the term saliency was used for discriminative features that help tracking a particular object by distinguishing it from the local background. 
These features were selected from predefined hand-crafted features (such as SIFT). 
In \cite{sonl}, the method extracts motion, appearance and location saliency maps and predicts the next location of objects via multiplication of all saliency maps with the input initial tracking bounding box(es). 
In \cite{scont}, color, texture and saliency features are used for mean-shift tracking algorithm. 
The previous works \cite{sdisc},\cite{sonl},\cite{scont} only use low-level visual saliency information which is opposite to our intention to exploit semantically higher level information. 
Moreover, in \cite{scont} the features are combined in a straightforward manner (concatenation). 
Compared to the prior studies, there are two key differences in our method: first, we exploit high-level deep salient object segmentation results.
Second, we propose a novel adaptive combination of saliency with other features to be tracked.

\section{Proposed Method}
\label{sec:format}

The proposed tracking performs two sequential steps at each frame, namely Correlation Filter Learning Step and Localization Step. We describe both steps next.

\subsection{Correlation Filter Learning Step}

A DCF based tracking algorithm is often initialized by a bounding box containing the object to be tracked.
The tracker first learns the appearance of the contents in this box. 
The tracker encodes this appearance by learning correlation filters that give the highest score with features extracted from an image region extending the target box. 
These correlation filters $g$ are multi-dimensional arrays that maximize the convolution $f*g$, where $f$ represents the features.
For example, ECO \cite{eco} tracker uses activations of conv1 and conv5 outputs of a VGG-m pretrained network \cite{vgg} and HoG \cite{hog} as features.
As indicated previously, these features may not be robust against appearance or viewpoint changes. 
Robustness against these conditions is best met at the semantically highest level, e.g. a classification/regression result of a deep network. 
For example, a person detector outputs 1 on the person or a salient object detector outputs 1 on the salient object regardless of any viewpoint/appearance change.
In this work we select this feature as the decision of a deep salient object detection network. 
The reason behind our selection is two-fold:

\begin{itemize}
\item Any other deep classification/regression task is category dependent (e.g. a typical semantic segmentation result covers 20 object classes only).
\item Deep salient object detection is usually much faster than for example deep semantic segmentation, yet it still encodes highly semantic information.
\end{itemize}

\begin{figure}[htb]

\centering
\centerline{\includegraphics[width=8.5cm]{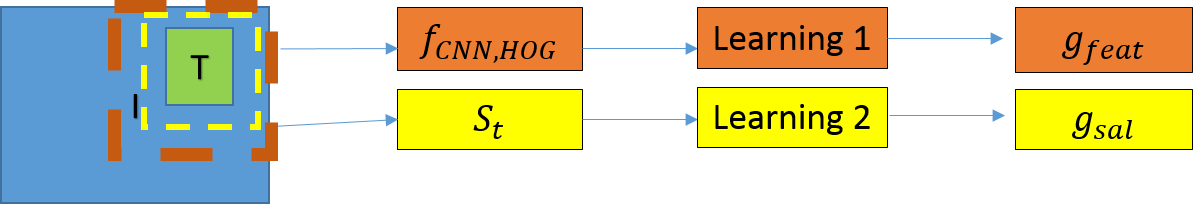}}

\caption{Filter Learning Step.}
\label{fig:filtlearn}
\end{figure}

Therefore, we add a deep salient object detection network output as an additional feature to ECO tracker framework.
We treat saliency and other features separately and learn their corresponding correlation filters separately.
The filter learning process is illustrated in Fig. \ref{fig:filtlearn}.

During filter learning step, the CNN and HoG features are extracted from a larger area (orange in Fig. \ref{fig:filtlearn}) than the target bounding box (green in Fig. \ref{fig:filtlearn}) since the appearance of the surrounding is also a discriminative factor for the target (at least for short-term correlation). 
However, saliency is a global, scale dependent feature and extracting saliency from a large region may result in less accurate detection of the salient object in the actual bounding box. 
For example, there might be other objects which are also salient in the enlarged window.
Therefore, we select the input region for the saliency feature extraction to be smaller than that of other features (yellow in Fig. \ref{fig:filtlearn}). 
We control this region's scale by another hand-tuned parameter. 
The correlation filters for saliency and other features are learned separately (see Fig. \ref{fig:filtlearn}). 
This shouldn't be confused with learning the features. 
The features are not learned, but extracted from pre-trained models.
In a DCF tracking method, the correlation filters are learned in a minimization framework where the cost is the Euclidean error between the ground truth and the convolution of the filter with the extracted feature. 
Ground truth is usually selected as a 2D Gaussian with a peak in the center of the object bounding box. 
In this paper, we follow the approach in ECO \cite{eco}, where the convolutions are utilized in continuous domain with periodic interpolation function and an additional term is added to the cost function to mitigate the periodic assumption. 
We also adopt the factorization of the convolutional filters which improves the computational efficiency of the tracker. 

\subsection{Target Localization Step}

A search space is selected from the new frame $t+1$, which is obtained by enlarging the target location from the previous frame $t$. The enlargement is made at different factors for saliency search and other features' search following the justification provided in the previous section. Features are extracted from these search spaces. The previously learned correlation filters are convolved with the extracted features in order to obtain responses (scores). Saliency scores and other feature scores are obtained separately and denoted as $R_{sal}$ and $R_{feat}$ respectively. 

Here, in order to combine saliency and other features, we propose to measure the reliability of the saliency map. 
This reliability is measured by the temporal similarity of the salient object detection. We measure the similarities of the saliency maps from the frames $t$ and $t+1$; we denote the vectorized saliency maps as $S_t$ and $S_{t+1}$, respectively. We choose the cosine similarity which is formulated as follows.

\begin{equation}
sim(S_t,S_{t+1})=\frac{S_t \cdot S_{t+1}}{{\|S_t\|}_2{\|S_{t+1}\|}_2}
\end{equation}

By construction, cosine similarity takes values in $[0,1]$. 

The importance (weight) of the saliency response at frame $t$ is denoted by $w(t)$ and is obtained as follows:

\begin{equation}
w(t+1)=K((1-\lambda)w(t)+\lambda sim(S_t,S_{t+1}))
\end{equation}

$K \in [0,1]$ denotes the maximum importance given on saliency, since the rest of the expression can at most take the value 1. $\lambda$ controls the contribution of the past importance and the new one and takes a value in $[0,1]$.

The final response map is then obtained as:

\begin{equation}
R=w(t+1)R_{sal}+R_{feat}
\end{equation}

\begin{figure*}[htb]

\centering
\centerline{\includegraphics[width=17cm]{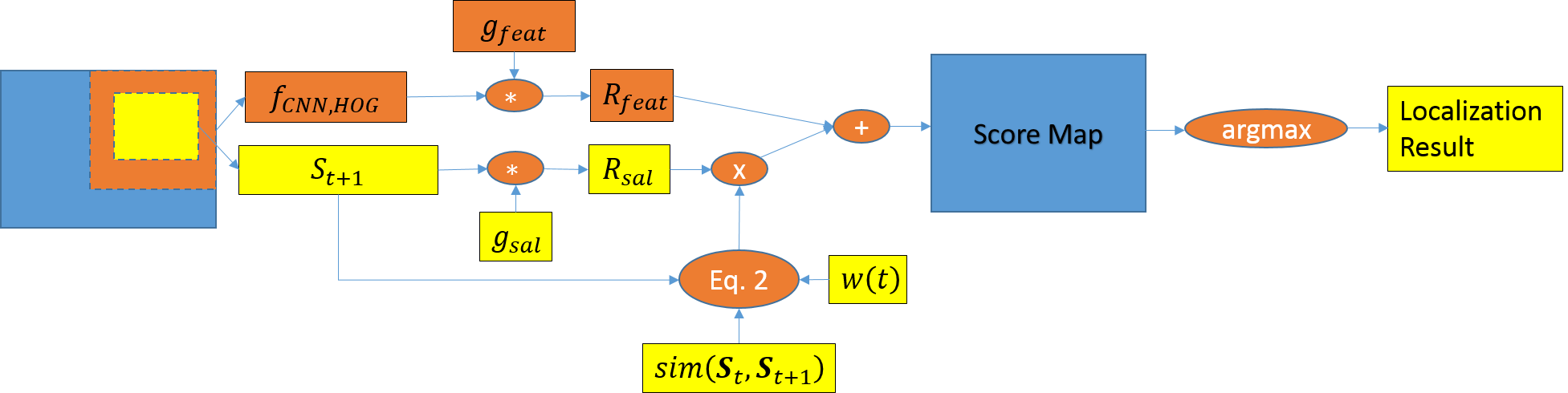}}

\caption{Target Localization Step.}
\label{fig:localize}
\end{figure*}

The target is localized in the coordinate whose value is largest in $R$. 
The target localization step is illustrated in Fig. \ref{fig:localize}.

\section{Experimental Results}
\label{sec:experiments}

In this section we provide quantitative evaluation of our method compared to the baseline method as well as some other state-of-the-art methods.

\begin{table*}
\caption{Comparison to the Baseline, OPE}
\label{table1}

\centering
\begin{subtable}{1\textwidth}
\centering
\resizebox{\textwidth}{!}{
    \begin{tabular}{|c|c|c|c|c|c|c|c|c|c|c|c|c|}
        \hline
         & IV & OPR & SV & OC & D & MB & FM & IPR & LR & BC & OV & ALL \\\hline
        ECOSAL & \textbf{0.701} & \textbf{0.712} & \textbf{0.709 }& \textbf{0.715} & \textbf{0.721} & \textbf{0.682} & \textbf{0.685} & \textbf{0.682} & \textbf{0.558} & \textbf{0.680} & \textbf{0.764 }& \textbf{0.715} \\\hline
        ECO & 0.699 & 0.710 & 0.702 &0.715 & 0.718 &0.671 & 0.672 & 0.678 & 0.513 & 0.674 & 0.726 & 0.711 \\\hline
    \end{tabular}}
\caption{Comparison to the Baseline - OTB-2013 dataset, One Pass Evaluation}
\end{subtable}

\begin{subtable}{1\textwidth}
\centering
\resizebox{\textwidth}{!}{
    \begin{tabular}{|c|c|c|c|c|c|c|c|c|c|c|c|c|c|}
        \hline
         & SV & AR & LR & FM & FO & PO & OV & BC & IV & VC & CM & SO & ALL \\\hline
        ECOSAL &\textbf{ 0.605} & \textbf{0.530} & 0.466 &\textbf{ 0.501 }& 0.344 & \textbf{0.554} & \textbf{0.524} & \textbf{0.491} & \textbf{0.562} & \textbf{0.575} &\textbf{ 0.626} &\textbf{ 0.647} &\textbf{0.648} \\\hline
        ECO & 0.595 & 0.503 & \textbf{0.469} & 0.487 &\textbf{ 0.363 }& 0.550 & 0.512 & 0.486 & 0.533 & 0.544 & 0.612 & 0.634 & 0.638 \\\hline
    \end{tabular}}

\caption{Comparison to the Baseline - UAV-123 dataset, One Pass Evaluation}
\end{subtable}

\end{table*}

\subsection{Implementation Details}

We empirically select the parameters of our method as follows: $\lambda=0.01, K=0.25, w(0)=0.125.$
Since we use saliency as a supplementary feature, we suppress its maximum importance to a quarter of other features by setting $K=0.25$. 
We initialize the saliency at an average measure which is the half of the maximum importance by setting $w(0)=0.125$. 
Finally, since the saliency map can be noisy and can focus on irrelevant areas outside the target window sometimes, that is why we update its importance very slowly by setting $\lambda=0.01$.
We use ECO \cite{eco} as the DCF based tracker due to its state-of-the-art performance and fast run-time.
We use DHS \cite{dhsnet} as the saliency extraction method due to its high performance and fast run-time.

\subsection{Datasets and Comparison Measures}

We use a widely used dataset OTB-2013 \cite{otb} for evaluation of our algorithm. We also evaluate our method on the recently introduced challenging dataset for UAV-123 tracking \cite{uav}. 

The evaluation metric for the tracking performance is chosen as the area under the success curve. Success curve is computed via plotting overlap threshold versus a success rate. The overlap between tracking result $T$ and the ground truth $GT$ is obtained as follows.

\begin{equation}
O(T,GT)=\frac{|T \cap GT|}{|T \cup GT|}
\end{equation}
If the overlap of $O(T,GT)$ for a frame is larger than a threshold, than that frame is successfully tracked. The success rate is the ratio of number of successfully tracked frames to the total number of frames in the video. By sweeping the overlap threshold, the success plot is obtained. The area under the curve (AUC) of success curve is a widely used evaluation metric.

There are three strategies for evaluation \cite{otb}. 
In one pass evaluation (OPE), the tracker is initialized in the first frame and and average success rate is reported. 
The tracking performance can be sensitive to (a) the spatial location of the bounding box (b) the frame on which the tracker is initialized. 
Therefore a spatial robustness evaluation (SRE) can be performed via starting the tracker at various positions. 
Similarly, temporal robustness evaluation (TRE) can be performed via initializing the tracker at various frames. 
For SRE, it is a common practice to use spatial shifts including 4 center shifts and 4 corner shifts, and 4 scale variations. 
Amount of shift is $10\%$ of target size, scale ratio varies among 0.8, 0.9, 1.1 and 1.2. 
Thus each tracker is evaluated 12 times for SRE and average performance is reported. 
For TRE, the video is divided into 20 segments, then the tracking is performed at each segment. Finally, the performance results are averaged.
For further robustness experiments, some videos in OTB-2013 are labeled according to challenges each video presents.

\begin{table*}

\caption{Robustness Analysis in OTB-2013 dataset}
\label{table2}
\centering
\begin{subtable}{1\textwidth}
\centering
\resizebox{\textwidth}{!}{
\begin{tabular}{|c|c|c|c|c|c|c|c|c|c|c|c|c|}
        \hline
         & IV & OPR & SV & OC & D & MB & FM & IPR & LR & BC & OV & ALL \\\hline
        ECOSAL & \textbf{0.607} &\textbf{ 0.637} & \textbf{0.634} &\textbf{ 0.657} & \textbf{0.642} & \textbf{0.618} & \textbf{0.621} & \textbf{0.601} & \textbf{0.496} & \textbf{0.60} & \textbf{0.681} & \textbf{0.647 } \\\hline
        ECO & 0.585 & 0.623 & 0.618 & 0.654 & 0.626 & 0.607 & 0.611 & 0.584 & 0.452 & 0.576 & 0.648 & 0.635 \\\hline
    \end{tabular}}
\caption{Comparison to the Baseline - OTB dataset, Spatial Robustness Estimation}
\end{subtable}

\begin{subtable}{1\textwidth}
\centering
\resizebox{\textwidth}{!}{
\begin{tabular}{|c|c|c|c|c|c|c|c|c|c|c|c|c|}
        \hline
         & IV & OPR & SV & OC & D & MB & FM & IPR & LR & BC & OV & ALL \\\hline
        ECOSAL & \textbf{0.672} & \textbf{0.690} & 0\textbf{.699} & \textbf{0.699} & \textbf{0.704} & \textbf{0.661} & \textbf{0.663} & \textbf{0.668} & \textbf{0.566} & \textbf{0.667} & \textbf{0.679} & \textbf{0.707}  \\\hline
        ECO & 0.664 & 0.683 & 0.685 & 0.695 & 0.701 & 0.651 & 0.649 & 0.659 & 0.536 & 0.657 & 0.670 & 0.700 \\\hline
    \end{tabular}}

\caption{Comparison to the Baseline - OTB dataset, Temporal Robustness Estimation}
\end{subtable}

\end{table*}

These challenges are: Illumination Variance (IV), Out of Plane Rotation (OPR), Scale Variance (SV), Occlusion (OC), Deformation (D), In Plane Rotation (IPR), Background Clutter (BC) and Out of View (OV). In UAV-123 dataset, additional labels are available such as Aspect Ratio Change (AR), Full Occlusion (FO), Partial Occlusion (PO), Camera Motion (CM), Similar Object (SO). It should be noted that not all videos are labeled in both datasets. It is a common practice to evaluate performance at each of these cases since it enables further evaluation of the robustness of the tracker to each case.

\subsection{Comparison to the Baseline}

Here, we compare our method ECOSAL with the baseline ECO \cite{eco}. In Table \ref{table1}, we show results for OTB and UAV datasets under OPE evaluation strategy. It can be observed that on average (ALL column in tables), our improved model ECOSAL performs superior to ECO in both datasets. The superior performance is more evident in the more challenging dataset UAV-123. In OTB-2013 dataset, ECOSAL is consistently better than ECO in all challenging cases except for occlusion cases (OC), where two algorithms perform the same. In UAV-123 dataset, ECOSAL performs better than ECO in 10 of 12 challenging cases. ECOSAL performs worse than ECO in only two cases: low resolution (LR) and full occlusion (FO) cases. From both datasets, we observe that the particular case of full occlusion is the only challenging case that our method performs inferior to the baseline ECO. In the fully occluded targets, the target might reappear far from the spatial extent of the final appearance. Therefore, because of the reduced spatial support for extraction and search of the saliency feature, our method might fail specifically in the full occlusion case if there is a large difference between the location of the target before and after the full occlusion. 
It should be noted that in Table \ref{table1}, the ALL column is not simply the average of the other entries in the table. This is because only some of the videos in the dataset are labeled as challenging cases.

In Table \ref{table2}, we compare spatial and temporal robustness estimations of our method ECOSAL with baseline ECO. We observe superior performance of our method in every single challenging case. This experiment shows that incorporation of saliency provides superior robustness to spatial and temporal variations during target initialization.

\subsection{Computational Complexity}

ECO runs at 11 fps on average in OTB-2013 dataset. 
Our improved model ECOSAL runs at 9.3 fps on average, adding a small computational burden over ECO.
Our experiments are conducted with GPU support, using Nvidia GTX 1080.

\subsection{Comparison to the State-of-the-Art}

We compare our methods to ECO \cite{eco}, MDNet \cite{mdnet}, SANet \cite{sanet}, MCPF \cite{mcpf} and CCOT \cite{ccot} on OTB-2013 dataset. The methods are selected according to their leading performance in benchmark \cite{bench}. As it is observed from the Table \ref{table3}, our method achieves better results than the state-of-the-art. ECO* and ECO denotes the evaluation of us and \cite{bench}'s accordingly. 

\begin{table}
\caption{Comparison to the State-of-the-Art in OTB-2013 dataset, One Pass Evaluation }
\label{table3}
\begin{adjustbox}{width=0.5\textwidth}
\begin{tabular}{|c|c|c|c|c|c|c|}
\hline
ECOSAL & ECO* & ECO & MDNet &SANet & MCPF & CCOT \\\hline
 \textbf{0.715} & 0.711 & 0.709 & 0.708 &0.686 & 0.677 & 0.672 \\\hline

\end{tabular}
\end{adjustbox}
\end{table}

%
%

\section{Conclusion}
\label{sec:conc}

We have proposed a method to incorporate deep salient object detection to the DCF based trackers. Our method automatically finds how to combine the saliency with other features, based on an adaptive measure of the confidence on the visual saliency. Our method has a robust performance, especially on challenging cases such as viewpoint and appearance changes. This is due to the semantically higher level information possessed by the salient object detection results, compared to the other features. Due to our method's adaptive combination strategy, it can suppress the temporally inconsistent salient object detection results.

\bibliographystyle{IEEE}
\bibliography{Template}

\end{document}